\documentclass[sigconf]{acmart}

\usepackage{makecell}

\usepackage{pifont}  

\usepackage{amssymb} 
\newcommand{\cmark}{\ding{51}}  
\newcommand{\xmark}{\ding{55}}  

\AtBeginDocument{%
  }

\setcopyright{acmlicensed}
\copyrightyear{2025}
\acmYear{2025}
\acmDOI{XXXXXXX.XXXXXXX}
\acmConference[Conference CIKM 2025]{Make sure to enter the correct
  conference title from your rights confirmation email}{June 03--05,
  2018}{Woodstock, NY}
\acmISBN{978-1-4503-XXXX-X/2018/06}

\begin{document}

\title{AGENTiGraph: A Multi-Agent Knowledge Graph Framework for Interactive, Domain-Specific LLM Chatbots}

\author{Xinjie Zhao}
\email{xinjie-zhao@g.ecc.u-tokyo.ac.jp}
\orcid{1234-5678-9012}
\affiliation{%
  \institution{The University of Tokyo}
  \country{Japan}
}

\author{Moritz Blum}
\email{mblum@techfak.uni-bielefeld.de}
\affiliation{%
  \institution{University of Bielefeld}
  \country{Germany}}

\author{Fan Gao}
\email{fangao0802@gmail.com}
\affiliation{%
  \institution{The University of Tokyo}
  \country{Japan}}

\author{Yingjian Chen}
\author{Boming Yang}
\email{yingjianchen@henu.edu.cn}
\email{boming.yang@weblab.t.u-tokyo.ac.jp}
\affiliation{%
  \institution{The University of Tokyo}
  \country{Japan}}

\author{Luis Marquez-Carpintero}
\author{Mónica Pina-Navarro}
\email{luis.marquez@ua.es}
\email{monica.pina@ua.es}
\affiliation{%
  \institution{University of Alicante}
  \country{Spain}}

\author{Yanran Fu}
\author{So Morikawa}
\email{fuyanran@stu.xmu.edu.cn}
\email{morikawa@civil.t.u-tokyo.ac.jp}
\affiliation{%
  \institution{The University of Tokyo}
  \country{Japan}}

\author{Yusuke Iwasawa}
\author{Yutaka Matsuo}
\email{iwasawa@weblab.t.u-tokyo.ac.jp}
\email{matsuo@weblab.t.u-tokyo.ac.jp}
\affiliation{%
  \institution{The University of Tokyo}
  \country{Japan}}

\author{Chanjun Park}
\email{chanjun.park@ssu.ac.kr}
\affiliation{%
  \institution{Soongsil University}
  \country{South Korea}}

\author{Irene Li}
\email{irene.li@weblab.t.u-tokyo.ac.jp}
\affiliation{%
  \institution{The University of Tokyo}
  \country{Japan}}

\renewcommand{\shortauthors}{Xinjie et al.}

\begin{abstract}
\textbf{AGENTiGraph} is a user-friendly, agent-driven system that enables intuitive interaction and management of domain-specific data through the manipulation of knowledge graphs in natural language. It gives non-technical users a complete, visual solution to incrementally build and refine their knowledge bases, allowing multi-round dialogues and dynamic updates without specialized query languages. The flexible design of AGENTiGraph, including intent classification, task planning, and automatic knowledge integration, ensures seamless reasoning between diverse tasks. Evaluated on a 3,500-query benchmark within an educational scenario, the system outperforms strong zero-shot baselines (achieving 95.12\% classification accuracy, 90.45\% execution success), indicating potential scalability to compliance-critical or multi-step queries in legal and medical domains, e.g., incorporating new statutes or research on the fly. Our open-source demo offers a powerful new paradigm for multi-turn enterprise knowledge management that bridges LLMs and structured graphs.
\end{abstract}


  


\maketitle

\section{Introduction}
Large Language Models (LLMs) have catalyzed a paradigm shift in knowledge-intensive applications~\cite{zhuang2023toolqadatasetllmquestion, gao-etal-2024-evaluating-large, ke2024enhancing, yang2023ascle}. However, they struggle with factual grounding, data provenance, and privacy-sensitive scenarios~\cite{gao-etal-2024-evaluating-large, xu2024knowledgeconflictsllmssurvey, augenstein2024factuality, yang2024retrieval}. In contrast, Knowledge Graphs (KGs) structurally encode entities and relations, providing a transparent, logically consistent framework for storing and querying domain-specific knowledge~\cite{nickel2015review, yang-etal-2024-kg, DBLP:conf/dl4kg/LiY23}. When harnessed in conjunction with LLMs, KGs have the potential to anchor language models in robust, auditable repositories of knowledge, thereby enhancing both accuracy and interpretability. Nevertheless, conventional query languages (e.g., SPARQL~\cite{Banerjee_2022}, Cypher~\cite{10.1145/3183713.3190657}) require technical expertise, limiting the accessibility for non-experts~\cite{10.3233/SW-160247, quamar2022NLInterfaces, castelltort2018handling, ji2021survey, sima2023sparql}. This limitation is especially critical in high-stakes fields like legal and medical domains, where users must construct proprietary knowledge bases, ensure privacy, control reasoning, and incorporate emerging information such as regulations and research~\cite{Tuggener2024SoYW}.

\begin{table}[t]
\centering
\resizebox{0.42\textwidth}{!}{
\begin{tabular}{lccc}
\toprule
\textbf{\makecell[l]{Functionality}} & 
\textbf{\makecell[c]{LLM-based\\Chatbots}} & 
\textbf{\makecell[c]{GraphRAG}} & 
\textbf{\makecell[c]{AgentiGraph\\(ours)}} \\
\midrule
\makecell[l]{Basic QA}                  & \cmark & \cmark   & \cmark \\
\makecell[l]{Multi-round QA}            & \cmark & \cmark   & \cmark \\
\makecell[l]{Multi-hop Reasoning}       & \xmark     & \cmark   & \cmark \\
\makecell[l]{Private Data}              & \xmark     & \cmark    & \cmark \\
\makecell[l]{\textbf{Visualization}}             & \xmark     & \xmark     & \cmark \\
\makecell[l]{\textbf{User Interaction}}          & \xmark     & \xmark     & \cmark \\
\makecell[l]{\textbf{Graph Edits}}               & \xmark     & \xmark     & \cmark \\
\makecell[l]{\textbf{Realtime Updates}}          & \xmark     & \xmark     & \cmark \\
\makecell[l]{\textbf{Automated Workflow}}        & \xmark     & \xmark     & \cmark \\
\bottomrule
\end{tabular}}
\caption{Comparison of core functionalities between the LLM-based Chatbots, GraphRAG, and AgentiGraph.}
\label{tab:comparison}
\vspace{-7mm}
\end{table}

In response to these requirements, we introduce \textbf{AGENTiGraph} (\textbf{A}daptive \textbf{G}eneral-purpose \textbf{E}ntities \textbf{N}avigated \textbf{T}hrough \textbf{I}nteraction), a versatile system that unites LLM capabilities with modular, multi-agent processes to facilitate end-to-end knowledge graph management. Unlike existing frameworks that treat KGs merely as static data sources for question answering, AGENTiGraph empowers users to actively curate, manipulate, and visualize their graphs via natural language dialogue. By orchestrating specialized agents for intent classification, graph updates, and continuous knowledge integration, it ensures a chain of knowledge can be both tracked and audited, addressing pressing challenges in privacy, compliance, and multi-step reasoning. Importantly, AGENTiGraph’s emphasis on user-centric design lowers the technical barrier to KG adoption, enabling professionals in law and healthcare to manage proprietary data stores without forfeiting performance or security.

\begin{figure*}[t]
  \centering
  \includegraphics[width=0.82\textwidth]{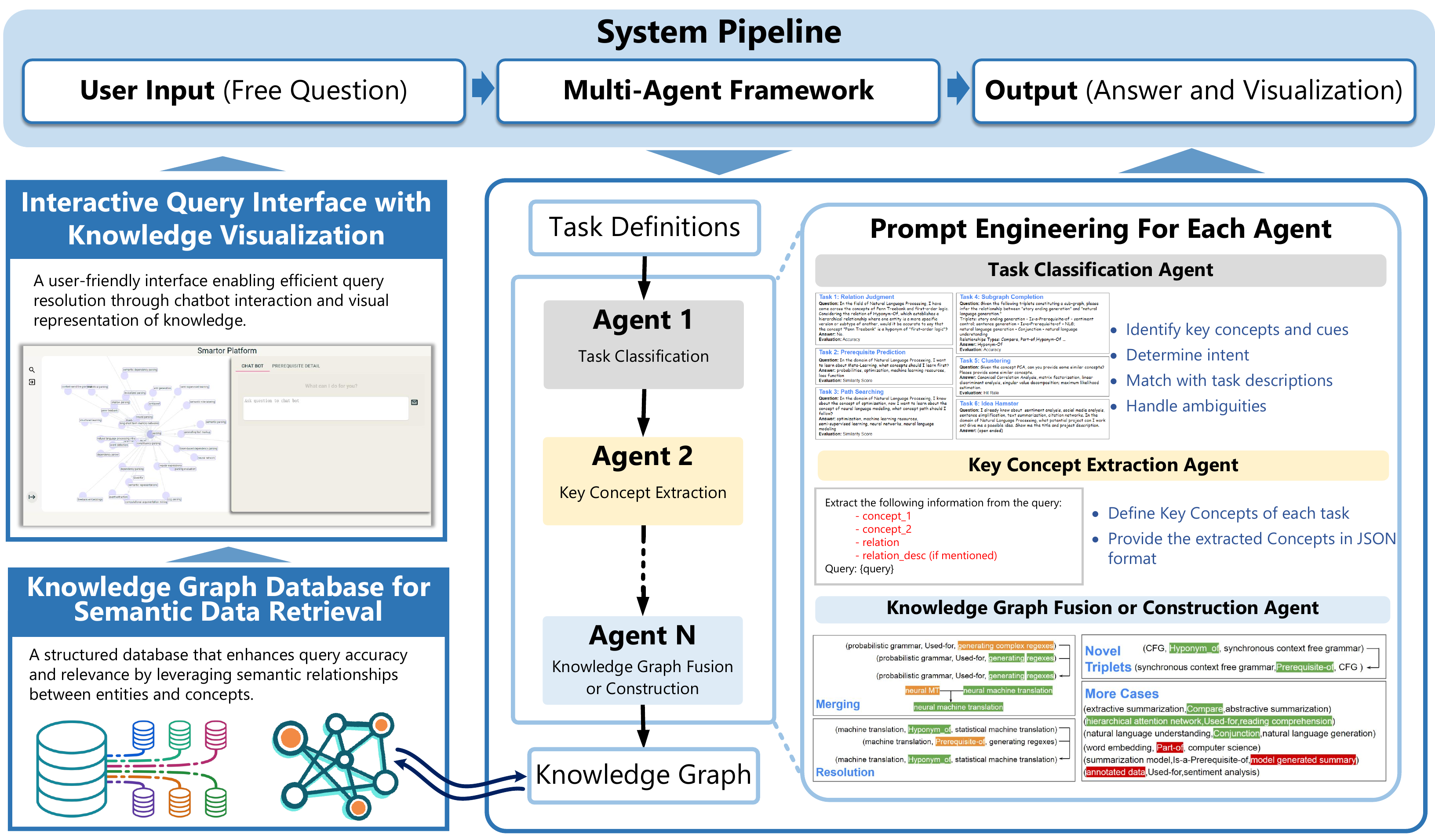}
  \caption{AGENTiGraph: A modular agent-based architecture for intelligent KG interaction and management. } 
  \label{fig:framework} 
  \vspace{-2mm}
\end{figure*}

AGENTiGraph is designed for cross-domain applicability, but in this work, we demonstrate its effectiveness through an educational scenario. On a 3,500-query benchmark, it achieves 95.12\% accuracy in user intent classification and a 90.45\% success rate in executing graph operations, outperforming state-of-the-art zero-shot baselines.
We summarize the principal contributions of this work as follows: (1) \textbf{Natural Language-Driven KG Interaction}: We introduce a modular architecture that enables users to explore and update knowledge graphs through intuitive natural language dialogues. Specialized agents for intent recognition, relation extraction, and real-time knowledge integration support transparent and auditable reasoning; 
(2) \textbf{Empirical and Dataset Contribution}: We extend TutorQA~\cite{yang2024graphusion} dataset to 3,500 queries, adding diverse free-form questions per task. AGENTiGraph outperforms state-of-the-art zero-shot baselines on this benchmark.
and (3) \textbf{Scalable, Privacy-Preserving Deployment}: We show how AGENTiGraph accommodates domain-specific constraints in legal and medical settings while dynamically incorporating new statutes, guidelines, and research\footnote{Demonstrated in our demo video. \textbf{Live Demo:} \url{https://drive.google.com/file/d/1IiA-XGveSgy1bw7d4ess_e8D6bQzA1P4/view?usp=sharing} \\
\textbf{Note on Availability:} 
Due to the high maintenance cost of keeping the API online, we cannot guarantee the chat-bot will always be functional. 
If you encounter server congestion or API delays, 
please consider deploying the system locally using the \textbf{Source Package}: \url{https://github.com/SinketsuZao/AGENTiGraph}.}
\section{AGENTiGraph Framework Design}
AGENTiGraph is designed to provide intuitive, seamless interaction between users and knowledge graphs $(G)$. It adopts a human-centric approach, allowing users to interact via natural language inputs $(q)$. To achieve this, we employ a pipeline of LLM-driven agents, each focused on a specific subtask. Each agent uses an LLM to interpret input, decompose it into actionable tasks, interact with the graph, and generate coherent responses $(a)$. This modular pipeline ensures the process remains flexible, interpretable, and extensible. Our pipeline contains the following workflow:

\noindent\textbf{1. User Intent Interpretation.}
The User Intent Agent interprets natural language input to determine the underlying intent $(i)$. Utilizing Few-Shot Learning \cite{10.1145/3386252} and Chain-of-Thought (CoT) reasoning \cite{10.5555/3600270.3602070}, it enables the LLM to handle diverse query types without extensive training data \cite{kwiatkowski2019natural}, ensuring adaptability to evolving needs.

\noindent\textbf{2. Key Concept Extraction.}
The Key Concept Extraction Agent performs Named Entity Recognition (NER) \cite{wang2020application} and Relation Extraction (RE) \cite{miwa-bansal-2016-end} on the input $(q)$. Guided by targeted examples, it maps extracted entities $(E)$ and relations $(R)$ to the knowledge graph via semantic similarity using BERT-derived vectors \cite{turton-etal-2021-deriving} to ensure accurate concept linking and efficiency.

\noindent\textbf{3. Task Planning.}
The Task Planning Agent decomposes the identified intent into a sequence of executable tasks $(T = {t_1, t_2, ..., t_n})$. Leveraging CoT reasoning, it models task dependencies, optimizes execution order, and generates structured sequences, particularly effective for complex queries requiring multi-step reasoning \cite{fu2023complexitybasedpromptingmultistepreasoning}.

\noindent\textbf{4. Knowledge Graph Interaction.}
The Knowledge Graph Interaction Agent bridges tasks and the graph by generating a formal query $(c_k)$ for each task $(t_k)$. Combining Few-Shot Learning with the ReAct framework \cite{yao2023reactsynergizingreasoningacting}, it enables dynamic query refinement based on intermediate results, adapting to diverse graph structures and query languages without extensive pre-training.

\noindent\textbf{5. Reasoning.}
The Reasoning Agent applies logical inference, leveraging the LLM’s contextual understanding and reasoning capabilities \cite{sun-etal-2024-determlr}. By framing reasoning as logical steps, it enables flexible inference across diverse tasks, bridging structured knowledge and natural language.

\noindent\textbf{6. Response Generation.}
The Response Generation Agent synthesizes processed information into coherent answers, using CoT, ReAct, and Few-Shot Learning to produce structured, contextually relevant outputs. This ensures responses are informative and aligned with the user's query.

\noindent\textbf{7. Dynamic Knowledge Integration.}
The Update Agent handles dynamic knowledge integration by adding new entities $(E_{\text{new}})$ and relationships $(R_{\text{new}})$ to $G$: $G \leftarrow G \cup {E_{\text{new}}, R_{\text{new}}}$. It interfaces directly with the Neo4j database, using LLM-generated Cypher queries to update the graph \cite{miller2013graph}.

\section{System Demonstration}

\subsection{User Interface}
\begin{figure*}[t]
\vspace{-2mm}
  \centering
\includegraphics[width=0.82\textwidth]{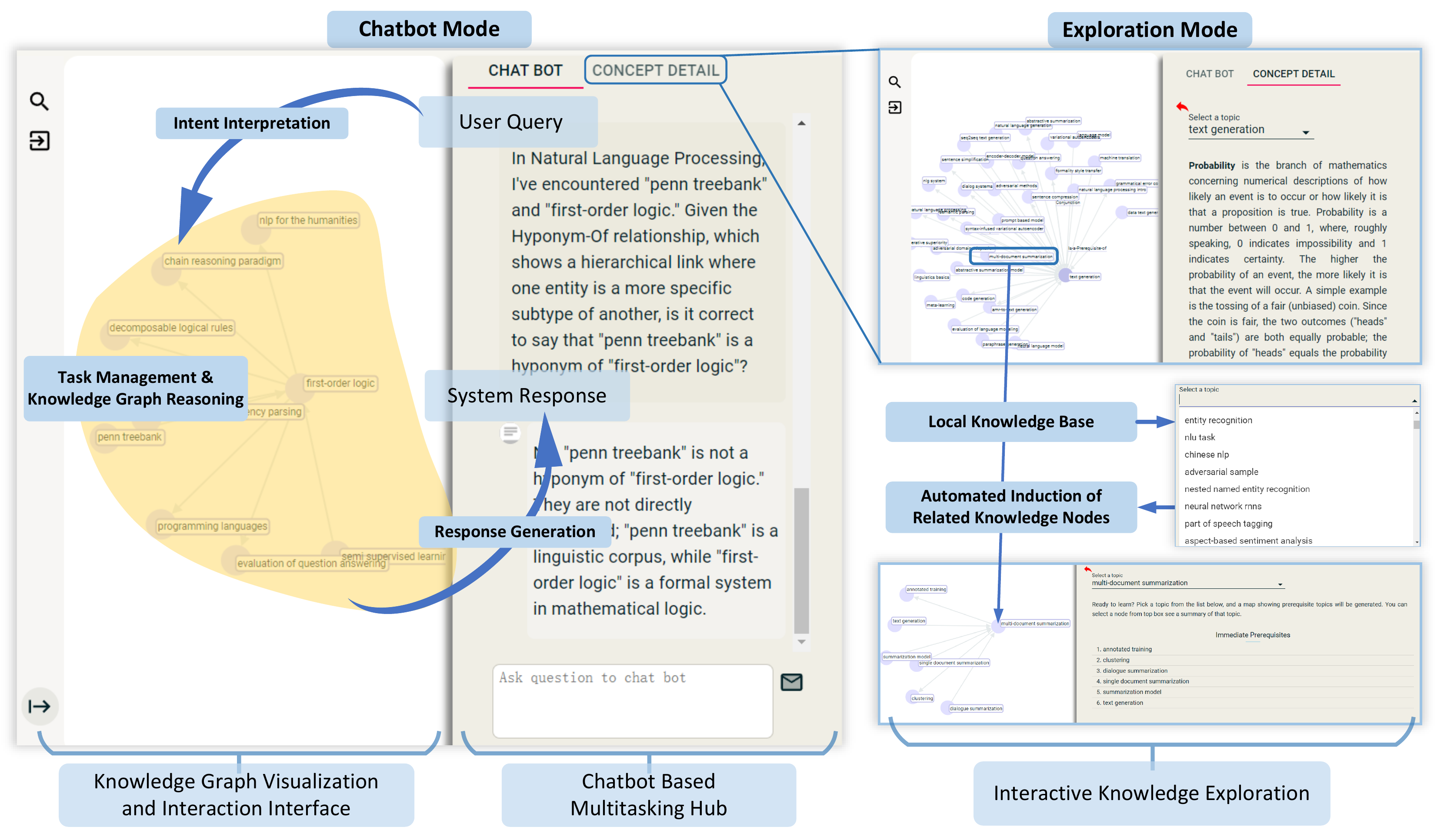}
\caption{Dual-Mode Interface Design: Conversational Interaction with Interactive Knowledge Exploration.}
\label{fig:ui_components}
\vspace{-2mm}
\end{figure*}

The AGENTiGraph interface is designed for intuitive use and efficient knowledge exploration, as shown in Figure~\ref{fig:ui_components}. It adopts a dual-mode interaction paradigm combining conversational AI with interactive knowledge navigation. The interface comprises three main components:
\textbf{Chatbot Mode} uses LLMs for intent interpretation and response generation via knowledge graph traversal, enabling nuanced natural language query processing.
\textbf{Exploration Mode} offers an interactive knowledge graph visualization with entity recognition, supporting hierarchy navigation and semantic relationship exploration.
\textbf{Knowledge Graph Management Layer} bridges the multi-agent system and the Neo4j database via the Bolt protocol, enabling efficient graph operations and retrieval.

\subsection{Task Design}
\label{sec:task_design}

To support user interaction with knowledge graphs and their diverse needs in knowledge exploration, AGENTiGraph provides a suite of pre-designed functionalities, inspired by the TutorQA, an expert-verified TutorQA benchmark, designed for graph reasoning and question-answering in the NLP domain~\cite{yang2024graphusion}. Specifically, AGENTiGraph supports the following tasks currently: \textbf{Relation Judgment} for verifying semantic connections; \textbf{Prerequisite Prediction} to identify foundational concepts; \textbf{Path Searching} for generating personalized learning paths; \textbf{Concept Clustering} to reveal macro-level knowledge structures; \textbf{Subgraph Completion} for uncovering hidden associations; and \textbf{Idea Hamster}, which supports practical idea generation based on structured knowledge.

AGENTiGraph's flexibility extends beyond predefined functionalities. Users can pose any question or request, and the system automatically determines whether it falls within the six categories. If not, it treats the input as a \textbf{free-form query}, employing a flexible approach to address specific needs. Users with specific requirements can also design custom agents or reconfigure existing ones to create tailored functionalities, ensuring AGENTiGraph evolves with diverse and changing user needs, and serves as a versatile platform for both guided and open-ended knowledge discovery.

\begin{table}[t]
\centering
\footnotesize
\resizebox{0.35\textwidth}{!}{
\begin{tabular}{lccc}
\toprule
\textbf{Model / Setting} & \textbf{Acc.} & \textbf{F1} & \textbf{Exec. Success} \\
\toprule
\multicolumn{4}{c}{\textit{\textbf{LLM Zero-shot}}} \\
LLaMa 3.1-8b & 0.6234 & 0.6112 & 0.5387 \\
LLaMa 3.1-70b & 0.6789 & 0.6935 & 0.5912 \\
Gemini-1.5 pro & 0.8256 & 0.8078 & 0.7434 \\
GPT-4 & 0.7845 & 0.7463 & 0.7123 \\
GPT-4o & 0.8334 & 0.8156 & 0.7712 \\
\midrule
\multicolumn{4}{c}{\textit{\textbf{Few-shot Prompting (Pure LLM)}}} \\
GPT-4 (few-shot) & 0.8532 & 0.8291 & 0.7805 \\
\midrule
\multicolumn{4}{c}{\textit{\textbf{BERT-based Classifier (Fine-tuned)}}} \\
BERT-classifier & 0.6150 & 0.5985 & - \\
\midrule
\multicolumn{4}{c}{\textit{\textbf{AGENTiGraph (ours)}}} \\
LLaMa 3.1-8b & 0.8356 & 0.8178 & 0.7230 \\
LLaMa 3.1-70b & 0.8789 & 0.8367 & 0.7967 \\
Gemini-1.5 pro & 0.9389 & 0.9323 & 0.8901 \\
GPT-4 & 0.9234 & 0.8912 & 0.8778 \\
GPT-4o & \textbf{0.9512} & \textbf{0.9467} & \textbf{0.9045} \\
\bottomrule
\end{tabular}}
\caption{Evaluation of task classification accuracy and execution success with additional baselines. BERT model \cite{devlin-etal-2019-bert}, LLaMa models \cite{grattafiori2024llama3herdmodels}, GPT-4 and GPT-4o \cite{openai2024gpt4technicalreport}.}
\label{tab:task_classification}
\vspace{-8mm}
\end{table}

\section{Evaluation}
\subsection{Experimental Setup}

We developed an expanded test set addressing the limitations of the original TutorQA dataset\footnote{\url{https://huggingface.co/datasets/li-lab/tutorqa}}~\cite{yang2024graphusion}, which comprises 3,500 cases, with 500 queries for each of six predefined tasks and 500 free-form queries (\S\ref{sec:task_design}). The dataset was created by using LLMs to mimic student questions \cite{liu-etal-2024-conversational}, with subsequent human verification ensuring quality and relevance, resulting in a diverse query set closely resembling real-world scenarios \cite{Extance2023ChatGPTHE}.

Our evaluation focuses on two aspects:
\textbf{Query Classification}: Assessing the system's ability to categorize user inputs into seven task types (six predefined plus free-form), measured by accuracy and F1.
\textbf{Task Execution}: Evaluating whether it can generate valid outputs for each query, measured by execution success.
To address fairness concerns, we introduce additional baselines beyond zero-shot scenarios, comparing AGENTiGraph to:
(1) A few-shot LLM baseline, with a small set of labeled examples for intent classification and prompting the LLM directly.
(2) A fine-tuned BERT-based classifier trained on 500 labeled queries.
These baselines confirm that performance gains arise not just from in-context learning but from our structured, multi-step reasoning and modular design.

\begin{table}[t]
\centering
\footnotesize
\resizebox{0.35\textwidth}{!}{
\begin{tabular}{lcc}
\toprule
\textbf{Aspect} & \textbf{Mean Rating (1-7)} \\
\midrule
Interface Intuitiveness & 5.8  \\
Response Comprehensibility & 6.0 \\
Relation Judgment Accuracy & 6.3 \\
Path Searching Clarity & 5.9  \\
Overall Satisfaction & 6.0  \\
\bottomrule
\end{tabular}}
\caption{Summary of user study results.}
\label{tab:user_study}
\vspace{-8mm}
\end{table}

\subsection{User Intent Identification \& Task Execution}

Table \ref{tab:task_classification} presents our experimental results. We first compared AGENTiGraph with zero-shot methods across multiple LLMs. To address concerns that our agent-based pipeline’s improvements might primarily stem from in-context learning, we introduced two additional baselines: a few-shot prompted GPT-4 and a fine-tuned BERT-classifier. The few-shot GPT-4 baseline demonstrates the effect of prompt engineering on performance, while the BERT-classifier offers a non-LLM, supervised perspective.\footnote{We attempted to use BERT for user intention modeling, but it performed poorly. As a result, we omit the execution success metric here.}

Our results show that AGENTiGraph still provides substantial gains over these new baselines. For instance, GPT-4o integrated with AGENTiGraph achieves a 95.12\% accuracy in task classification, which highlights that AGENTiGraph’s hierarchical, multi-step reasoning pipeline and structured approach—beyond just zero-shot or few-shot prompting—drives meaningful improvements.
These improvements are consistent across all model sizes, even for the simpler LLaMa 3.1-8b, suggesting that the agent-based pipeline amplifies the capabilities of underlying models. While the performance gap between zero-shot and AGENTiGraph narrows for larger models, AGENTiGraph’s approach remains robust, indicating that our framework’s advantages stem from its method of orchestrating the agents and processes rather than model size. The gap between classification accuracy and execution success persists, reflecting a complex interplay between understanding the user’s intent and executing the corresponding tasks. Yet, AGENTiGraph narrows this gap more effectively than the baselines, suggesting that multi-step task planning and reasoning agents help bridge the understanding-execution divide.

\subsection{System Usability and User Feedback}

Participants interacted with AGENTiGraph and rated various aspects on a 7-point Likert scale. We summarize key findings in Table~\ref{tab:user_study}, where users generally found the interface intuitive (mean ratings around 5.8), the responses comprehensible (mean around 6.0), and the system effective for relation judgment tasks (mean 6.3). While path-searching tasks received slightly lower scores (mean 5.9) due to requests for more visual detail, overall satisfaction remained high at about 6.0. Compared with a baseline system (ChatGPT-4o), 64\% of the queries were rated as more concise and contextually focused with AGENTiGraph. About 10\% of queries highlighted a need for more detailed explanations, especially for complex tasks. 

\section{Conclusion}
AGENTiGraph presents a novel approach to knowledge graph interaction, leveraging an adaptive multi-agent system to bridge LLMs and knowledge representations. Our platform outperforms existing solutions in task classification and execution, and is particularly suited to high-privacy requirements in areas such as legal and healthcare, demonstrating potential to revolutionize knowledge management across domains.


\bibliographystyle{ACM-Reference-Format}
\bibliography{sample-base}










\end{document}